%% file: iclr2026_conference.tex
\documentclass{article} % For LaTeX2e
\usepackage{iclr2026_conference,times}

\input{math_commands.tex}

\usepackage{xcolor}
\usepackage{pgfplots}
\usepackage{hyperref}
\usepackage{url}
\usepackage{booktabs}
\usepackage{graphicx}
\usepackage[table]{xcolor}

\usepackage{booktabs}
\usepackage{tabularx}
\usepackage{xcolor}
\usepackage{pgfplots}
\usepgfplotslibrary{groupplots}
\definecolor{color2}{RGB}{98,190,166}
\definecolor{color3}{RGB}{253,186,107}
\definecolor{color4}{RGB}{235,96,70}
\definecolor{color5}{RGB}{111,146,225}
\definecolor{color6}{RGB}{175,135,220}
\definecolor{opdblue}{RGB}{45,92,185}       % OPD-KL: 深蓝色
\definecolor{ctsred}{RGB}{220,62,48}         % CTS: 深红色
\definecolor{expertgreen}{RGB}{0,135,100}    % Expert: 深绿色
\definecolor{basegray}{RGB}{65,65,65}        % Base: 深灰色
\usepackage{graphicx}
\usepackage{caption}
\usepackage{enumitem}
\newcommand{\method}{\textsc{FlowCTS-OPD}}
\newcommand{\methodoff}{\textsc{FlowCTS-SFT}}
\pgfplotsset{
    iclrfigure/.style={
        tick label style={font=\footnotesize},
        label style={font=\footnotesize},
        axis line style={line width=0.55pt},
        tick style={line width=0.50pt},
        scaled ticks=false
    },
    opdcurve/.style={
        color=opdblue,
        line width=1.20pt,
        mark=o,
        mark size=1.10pt,
        mark options={
            solid,
            fill=white,
            draw=opdblue,
            line width=0.65pt
        }
    },
    ctscurve/.style={
        color=ctsred,
        line width=1.20pt,
        mark=square,
        mark size=1.05pt,
        mark options={
            solid,
            fill=white,
            draw=ctsred,
            line width=0.65pt
        }
    },
    baseline/.style={
        color=basegray,
        densely dotted,
        line width=1.05pt
    },
    expertline/.style={
        color=expertgreen,
        dashed,
        line width=1.05pt
    }
}
\pgfplotsset{compat=1.18}

\definecolor{plotblue}{HTML}{0072B2}
\definecolor{plotorange}{HTML}{D55E00}

\title{FlowCTS: On-policy Continuous Trajectory Supervision of Flow Models}

\iclrfinalcopy

\author{Kaiyang Ye$^{1}$\thanks{Equal contribution.}~~, Yuan Ge$^{1*}$, Junxiang Zhang$^1$, Bei Li$^1$, Ziming Zhu$^1$, Haishu Zhao$^1$, \\   
\textbf{Xiaoqian Liu$^{1}$, Chenglong Wang$^{1,2}$, Jingbo Zhu$^{1,2}$, Zhengtao Yu$^3$, Tong Xiao$^{1,2}$}\thanks{Corresponding author.} \\
$^1$ Northeastern University 
$^2$ NiuTrans Research 
$^3$ Kunming University of Science and Technology
}

\begin{document}

\maketitle
\fancyhead{}
\lhead{Preprint}
\begin{abstract}
% Flow matching models have become a powerful paradigm for text-to-image synthesis, yet effective post-training remains challenging: supervised fine-tuning uses fixed off-policy states, whereas reinforcement learning typically relies on sparse rewards. On-policy distillation provides dense reference supervision at student-visited states, however, how to extend OPD to flow models remains unexplored. 
%On-policy distillation (OPD) has emerged as an effective paradigm for large language model post-training, addressing the challenges of sparse rewards and exposure bias through dense reference supervision along student-generated trajectories. Flow models also face these challenges during post-training. However, how to extend OPD to flow models remains underexplored.
While on-policy distillation (OPD) effectively addresses sparse rewards and exposure bias in large language model post-training, its extension to flow models remains underexplored.
To this end, we propose Flow Continuous Trajectory Supervision (\textsc{FlowCTS}), which matches subsequent student and reference trajectories initialized from the same student-visited state. Using the integral relation between trajectories and velocity fields, we derive a temporally weighted velocity-matching upper bound and discretize it into practical objectives parameterized by the number of supervision steps. Under a multi-reference setup, single-state \textsc{FlowCTS-OPD} outperforms vanilla KL-based OPD with faster convergence. \textsc{FlowCTS-OPD} improves GenEval from $0.90$ to $0.93$, OCR from $0.90$ to $0.92$, and PickScore from $22.75$ to $23.06$, while outperforming a mixed-reward RL baseline across all target metrics. Further analysis reveals a clear temporal supervision mismatch in vanilla KL-based OPD arising from its auxiliary SDE transition kernels. Beyond on-policy setting, \textsc{FlowCTS} also consistently outperforms vanilla SFT , particularly on OCR, while increasing supervision steps exhibit a trade-off between richer trajectory information and greater optimization difficulty.
\end{abstract}

\begin{figure*}[h]
    \centering
    \includegraphics[width=\textwidth]{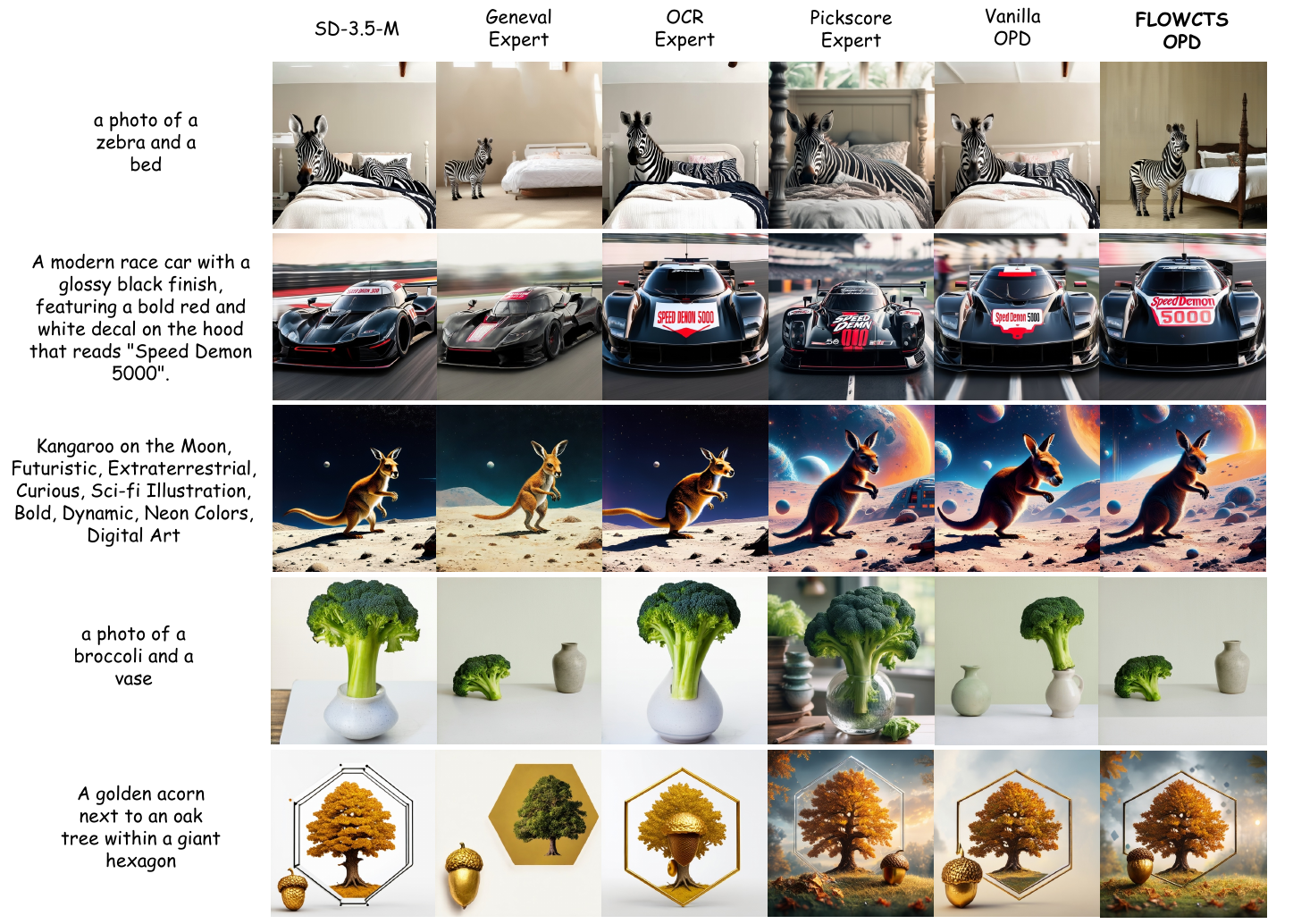}
    \caption{
    Qualitative comparison of generated images.
    Our method \textsc{FlowCTS-OPD} produces images with improved alignment and visual quality compared with previous approaches.
    }
    \label{fig:qualitative}
\end{figure*}

\section{Introduction}

Qwen3, MiMo-v2, and GLM-5 have recently attributed part of their post-training gains to an emerging paradigm: on-policy distillation (OPD)~\citep{yang2025qwen3,xiao2026mimo,zeng2026glm}. 
This approach, as illustrated in Figure \ref{fig:main} (a), combines on-policy training with dense reference supervision ~\citep{agarwal2024policy,gu2024minillm}. 
Although its training dynamics remain incompletely understood, these successes highlight the value of OPD for future generative-model development~\citep{li2026rethinking,hou2026uni}. 
In imitation learning, it is well established that learned policies benefit from training under their induced state distributions, rather than under those encountered by the expert~\citep{ross2010efficient,ross2011reduction}. This facilitates recovery from mistakes and helps reduce compounding errors. 
Moreover, OPD provides such benefits by training on the student’s rollout distribution with dense token-level reference supervision, thereby reducing reliance on sparse outcome rewards~\citep{lu2025onpolicydistillation,ma2026mopd,yang2026learning}. 
Inspired by these, this work investigates how to effectively formulate dense OPD supervision for flow models.

To this end, we first revisit the form of supervision in token-based autoregressive language models~\citep{gu2024minillm,li2026rethinking}, whose original objective minimizes the sequence-level reverse KL divergence between the student's and the reference's output sequence distributions given a prompt. 
The chain-rule factorization of language modeling decomposes this objective exactly into \textit{next token supervision} over a finite vocabulary at student-generated prefixes~\citep{agarwal2024policy,wen2023f}. 
However, flow models parameterize a continuous velocity field to model the continuous state trajectory from noise to data and lack a natural next-step conditional distribution~\citep{lipman2022flow,liu2022flow,tong2023improving,liu2023instaflow,kornilov2024optimal,frans2025one,geng2026mean}. This motivates us to revisit the essence of \textit{sequence supervision} in OPD, shifting the supervision target from next-step distributions to the trajectory itself.

We therefore propose Flow Continuous Trajectory Supervision (\textsc{FlowCTS}), which supervises the discrepancy between the student and reference continuous trajectories that evolve from the same shared state. \method~ instantiates this objective in the on-policy setting, where the shared starting state is sampled from the current student's rollout. We further derive an upper bound of the continuous trajectory supervision as a tractable optimization objective and discretize it into a finite number of optimization steps. The discretized objective is parameterized by the number of supervision steps $K$, covering single-state velocity matching and multi-step trajectory supervision. In the on-policy setting, single-state \method~ outperforms vanilla KL-based OPD with faster convergence. Specifically, it improves GenEval from $0.90$ to $0.93$, OCR from $0.90$ to $0.92$, and PickScore from $22.75$ to $23.06$, while also showing improved out-of-distribution generalization. Qualitative examples are shown in Figure \ref{fig:qualitative}. Further analysis reveals a clear temporal supervision mismatch in vanilla KL-based Flow-OPD, which constructs transition kernels via an auxiliary SDE. 

More broadly, \textsc{FlowCTS} does not restrict how the shared starting state is sampled. When the starting state is constructed from an off-policy data trajectory, the same formulation extends to \textsc{FlowCTS-SFT}. Experiments show that \textsc{FlowCTS-SFT} consistently outperforms vanilla SFT in overall score, with the most pronounced gains on OCR. Although increasing the supervision steps does not bring consistent improvements, \textsc{FlowCTS} provides a new scaling law dimension of flow models that trades richer trajectory information against greater optimization difficulty. In summary, our contributions are:

\begin{itemize}
    
    \item We take an early exploration toward dense on-policy supervision for flow models. Revisiting the essence of sequence supervision in OPD for language modeling, we propose \textsc{FlowCTS} from a continuous trajectory supervision perspective.
    
    \item The discretized objective of \textsc{FlowCTS} covers both single-state and multi-step supervision, introducing the supervision step as a new scaling dimension for flow models.
    
    \item We extended \textsc{FlowCTS} to both on-policy distillation and off-policy SFT settings, with experiments validating its effectiveness in both settings.
    
    \item We attribute the advantage of \textsc{FlowCTS}  to the dense supervision in early denoising stages, where the student–reference discrepancy is largest. In contrast, KL-based FlowOPD concentrates supervision in late stages and suffers from this temporal supervision mismatch.
\end{itemize}

\section{Preliminaries}
In this section, we begin by reviewing flow matching and standard on-policy distillation, which together motivate our trajectory-based OPD formulation.
\paragraph{Flow Matching.}
Flow matching defines a transport trajectory between the data and noise distributions via an ordinary differential equation (ODE). Formally, let $\vx_0 \sim p_{\mathrm{data}}$ be a data sample and $\vx_1 \sim p_{\mathrm{noise}}=\mathcal{N}(\vzero,\mI)$ be a noise sample. The transport trajectory is commonly defined by linear interpolation~\citep{liu2022flow}:
\begin{equation}
\vx_t = (1 - t)\vx_0 + t\vx_1, \quad t \in [0,1].
\end{equation}
The trajectory satisfies the ODE $d\vx_t = \vv_t(\vx_t, t)dt$, and under the linear path, the conditional target velocity is $\vu_t(\vx_t \mid \vx_0, \vx_1) = d\vx_t/dt = \vx_1 - \vx_0$. Flow matching trains a neural velocity field $\vv_{\vtheta}(\vx_t, t)$ by minimizing
\begin{equation}
\mathcal{L}_{\mathrm{FM}}(\vtheta)
=
\E_{t, \vx_0, \vx_1}
\left[
\left\lVert
\vv_{\vtheta}(\vx_t, t)
-
(\vx_1 - \vx_0)
\right\rVert^2
\right].
\end{equation}

\paragraph{On-Policy Distillation.}
OPD provides reference supervision on samples generated by the current student. 
Given a prompt $x \sim \mathcal{D}_x$, the student samples a response $\hat{y}_{1:T}\sim\pi_{\vtheta}(\cdot\mid x)$. 
At each student-generated prefix $(x,\hat{y}_{<t})$, the student and reference define next-token distributions
$p_{\vtheta}^t(\cdot)=\pi_{\vtheta}(\cdot\mid x,\hat{y}_{<t})$ and
$p_{\mathrm{T}}^t(\cdot)=\pi_{\mathrm{T}}(\cdot\mid x,\hat{y}_{<t})$.

A standard OPD objective minimizes the sequence-level reverse KL between the student and reference:
\begin{equation}
\mathcal{L}_{\mathrm{OPD}}(\vtheta)
=
\E_{x\sim\mathcal{D}_x}
\left[
\KL\left(
\pi_{\vtheta}(\cdot\mid x)
\,\Vert\,
\pi_{\mathrm{T}}(\cdot\mid x)
\right)
\right].
\end{equation}
By autoregressive factorization, this objective can be written as the sum of token-level KL terms on student-generated prefixes.
\begin{equation}
\mathcal{L}_{\mathrm{OPD}}(\vtheta)
=
\E_{x\sim\mathcal{D}_x,\ \hat{y}_{1:T}\sim\pi_{\vtheta}(\cdot\mid x)}
\left[
\sum_{t=1}^{T}
\KL\left(
p_{\vtheta}^t
\,\Vert\,
p_{\mathrm{T}}^t
\right)
\right].
\end{equation}
Unlike autoregressive models, whose sequence factorization naturally yields next-token conditional distributions for local matching, flow models generate samples through a continuous state trajectory from noise to data and do not natively provide such a next-step distribution. Yet, given the same visited state, the student and reference models define distinct subsequent trajectories. This observation motivates us to formulate OPD at the trajectory level by matching their subsequent evolution over a finite time interval.

\section{A Continuous-Trajectory Perspective on OPD for Flow Models}
\begin{figure*}[t]
    \centering
    \includegraphics[
        width=\linewidth,
        trim={0 2.5cm 0 0},
        clip
    ]{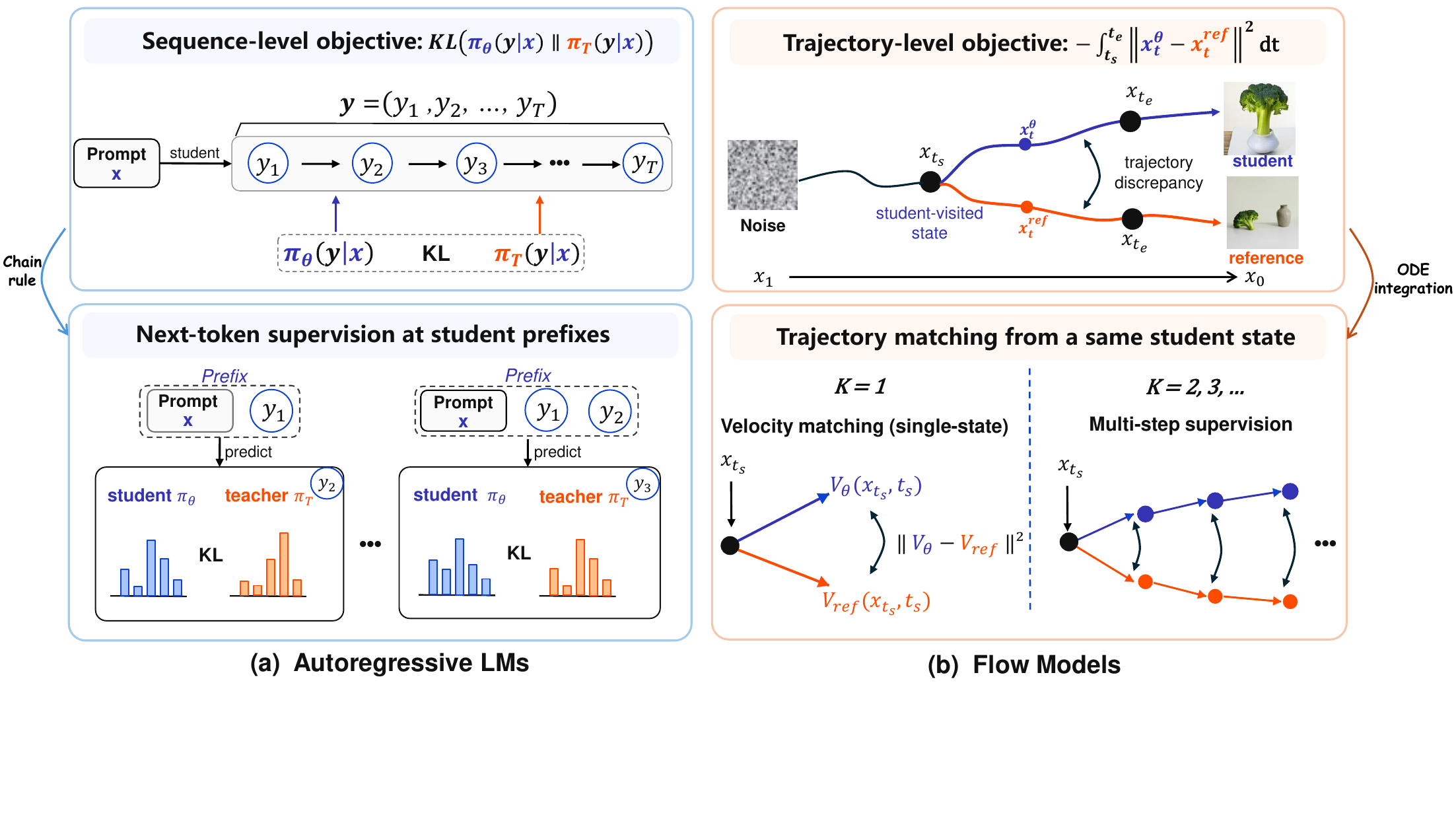}
    \caption{
    \textbf{Autoregressive OPD and trajectory-level supervision for flow models.} In autoregressive models, sequence-level reverse KL  decomposes by the chain rule into next-token supervision at student-generated prefixes. Since flow models do not naturally admit an analogous next-step factorization, FlowCTS instead matches student and reference trajectories initialized from the same student-visited state. Using the integral relation between trajectories and velocity fields, we derive a temporally weighted velocity-matching upper bound and discretize it into practical objectives parameterized by the supervision steps K, with K=1 recovering single-state velocity matching.
    }
    \label{fig:main}
\end{figure*}

In this section, we study how to formulate OPD for flow matching models from a continuous-trajectory perspective, as illustrated in
Figure~\ref{fig:main}. Under OPD, we first sample a student-visited state $\vx_{t}$ along the current student's generation trajectory. From this state, the student and reference velocity fields induce two subsequent trajectories. We formulate their discrepancy, derive a tractable velocity-based upper bound, and discretize it into a family of practical training objectives. We further discuss how the same formulation extends to off-policy starting states.

\subsection{A Trajectory-Level Formulation}

We begin with a trajectory-level formulation over a reverse-time
segment. Let $0 \leq t_{\mathrm{e}} < t_{\mathrm{s}} \leq 1$, where $t_{\mathrm{s}}$ and $t_{\mathrm{e}}$ denote the start and end times of the segment, respectively. 
Let $\Phi_{t_{\mathrm{s}}\to t}^{\vtheta}$ and
$\Phi_{t_{\mathrm{s}}\to t}^{\mathrm{ref}}$ denote the flow maps induced by the student and reference velocity fields, respectively. Given a starting state  $\vx_{t_{\mathrm{s}}}$ and a condition $c$, the corresponding trajectories are $\vx_t^{\vtheta}=\Phi_{t_{\mathrm{s}}\to t}^{\vtheta}
(\vx_{t_{\mathrm{s}}},c)$ and
$\vx_t^{\mathrm{ref}}=\Phi_{t_{\mathrm{s}}\to t}^{\mathrm{ref}} (\vx_{t_{\mathrm{s}}},c)$. Their discrepancy is defined as:

\begin{equation}
\mathcal{L}_{\mathrm{CTS}}(\vtheta)
=
\E_{\vx_{t_{\mathrm{s}}},c}
\left[
\int_{t_{\mathrm{e}}}^{t_{\mathrm{s}}}
\left\lVert
\vx_t^{\vtheta}
-
\vx_t^{\mathrm{ref}}
\right\rVert^2
dt
\right].
\end{equation}

We refer to this trajectory-level formulation as Continuous Trajectory
Supervision (CTS). It defines an on-policy distillation objective when the starting state $\vx_{t_{\mathrm{s}}}$ is sampled from the current student rollout, and an off-policy supervision objective when it is constructed from an offline data trajectory.

\subsection{A Velocity-Based Upper Bound}
\label{sec:method}
Flow trajectories are generated by integrating time-dependent velocity
fields~\citep{lipman2022flow,liu2022flow}. Therefore, the discrepancy between the student and reference trajectories is determined by the accumulated mismatch between their velocities. Based on this relation, we derive a weighted velocity-matching upper bound of the trajectory loss, yielding a tractable objective defined directly on the model outputs.

Specifically, since both trajectories start from the same state $\vx_{t_{\mathrm{s}}}$, their states at any $t\in[t_{\mathrm{e}},t_{\mathrm{s}}]$ are obtained by integrating the corresponding velocity fields:
\begin{equation}
\vx_t^{\vtheta}
=
\vx_{t_{\mathrm{s}}}
-
\int_t^{t_{\mathrm{s}}}
\vv_{\vtheta}(\vx_r^{\vtheta},r,c)\,dr,
\qquad
\vx_t^{\mathrm{ref}}
=
\vx_{t_{\mathrm{s}}}
-
\int_t^{t_{\mathrm{s}}}
\vv_{\mathrm{ref}}(\vx_r^{\mathrm{ref}},r,c)\,dr.
\end{equation}

Subtracting the two equations gives:
\begin{equation}
\vx_t^{\vtheta}
-
\vx_t^{\mathrm{ref}}
=
-
\int_t^{t_{\mathrm{s}}}
\left[
\vv_{\vtheta}(\vx_r^{\vtheta},r,c)
-
\vv_{\mathrm{ref}}(\vx_r^{\mathrm{ref}},r,c)
\right]dr.
\end{equation}

This identity shows that the trajectory discrepancy arises from the
accumulated velocity mismatch along the integration path. It leads to the
following weighted velocity-matching upper bound, which we term the velocity
upper bound (VUB):
\begin{equation}
\mathcal{L}_{\mathrm{CTS}}(\vtheta)
\leq
\mathcal{L}_{\mathrm{VUB}}(\vtheta)
=
\E_{\vx_{t_{\mathrm{s}}},c}
\left[
\int_{t_{\mathrm{e}}}^{t_{\mathrm{s}}}
w(r)
\left\lVert
\vv_{\vtheta}(\vx_r^{\vtheta},r,c)
-
\vv_{\mathrm{ref}}(\vx_r^{\mathrm{ref}},r,c)
\right\rVert^2
dr
\right],
\end{equation}
where the trajectory-derived weight is
$w(r)=\int_{t_{\mathrm{e}}}^{r}(t_{\mathrm{s}}-t)\,dt
=\big[(t_{\mathrm{s}}-t_{\mathrm{e}})^2-(t_{\mathrm{s}}-r)^2\big]/2$.
This weight arises from the trajectory bound rather than manual design.
It increases from zero at the segment end $t_{\mathrm{e}}$ to its maximum at the segment start $t_{\mathrm{s}}$, assigning larger weights to velocity errors closer to $t_{\mathrm{s}}$, as they affect a longer subsequent trajectory. A full derivation is provided in
Appendix~\ref{app:upper_bound}.

\subsection{Discrete Training Objective}

We now discretize the reverse-time segment $[t_{\mathrm{e}}, t_{\mathrm{s}}]$ into $K$ Euler steps. 
Let $h=(t_{\mathrm{s}}-t_{\mathrm{e}})/K$ and define $t_i=t_{\mathrm{s}}-ih$ for $i=0,\dots,K$. 
Starting from the same state $\vx_0^{\vtheta}=\vx_0^{\mathrm{ref}}=\vx_{t_{\mathrm{s}}}$, the student and reference rollouts are given by
\begin{equation}
\vx_{i+1}^{\vtheta}
=
\vx_i^{\vtheta}
-
h\vv_{\vtheta}(\vx_i^{\vtheta},t_i,c),
\qquad
\vx_{i+1}^{\mathrm{ref}}
=
\vx_i^{\mathrm{ref}}
-
h\vv_{\mathrm{ref}}(\vx_i^{\mathrm{ref}},t_i,c),
\end{equation}
for $i=0,\ldots,K-1$. Applying the same error-accumulation argument to the discretized trajectories yields the following practical objective:
\begin{equation}
\mathcal{L}_{\mathrm{CTS}}^{(K)}(\vtheta)
=
\E\left[
\sum_{i=0}^{K-1}
\alpha_i
\left\|
\vv_{\vtheta}(\vx_i^{\vtheta},t_i,c)
-
\vv_{\mathrm{ref}}(\vx_i^{\mathrm{ref}},t_i,c)
\right\|^2
\right],
\label{eq:discrete_cts}
\end{equation}

where $\alpha_i=[K(K+1)-i(i+1)]/2$ for
$i=0,\ldots,K-1$ is the trajectory-derived discrete weight. It reflects
how the velocity mismatch at step $i$ accumulates over the subsequent
trajectory. We present the weights in unnormalized form to make their
relative weighting explicit and normalize them to sum to one in
implementation. See Appendix~\ref{app:discrete_objective} for the full
derivation.

When the starting state is sampled from the current student's rollout, the resulting formulation defines a family of OPD objectives parameterized by the supervision step $K$. For $K=1$, we have $\alpha_0=1$, and the objective reduces to velocity matching at a single student-visited state. For $K>1$, supervision extends to multiple subsequent states along the student and reference trajectories. Thus, $K$ explicitly controls how far supervision extends along the trajectory, and we study its effect empirically in Section~\ref{sec:horizon}. When the starting state is constructed from an off-policy data trajectory, the same formulation yields the corresponding off-policy supervision objective.

\section{Experiments}
% Requires:
% \usepackage{booktabs}

\subsection{Experimental Setup}
\label{sec:experimental_setup}

\paragraph{On-policy setting.}
We evaluate FlowCTS in a multi-reference OPD setting based on SD3.5-Medium\citep{esser2024scaling}. We consider GenEval, OCR, and PickScore as target capabilities and use the corresponding officially released task-specific Flow-GRPO checkpoints as references~\citep{liu2026flow}, without additional models. For each mini-batch, prompts are sampled from a target-specific prompt set and routed to the corresponding reference model, while student-visited states are generated using the SDE rollout procedure adopted in Flow-GRPO. All OPD variants share the same prompt mixture, rollout procedure, reference routing, optimization hyperparameters, and inference configuration. We compare SDE and ODE rollouts in Appendix~\ref{app:sde_ode}
% Additional training details are provided in Appendix~\ref{app:training_details}.

\paragraph{Off-policy SFT setting.}
We further evaluate FlowCTS in an off-policy setting. Starting states and target trajectories are constructed from 50K image--text pairs sampled from Fine-T2I~\citep{ma2026fine}. We use SD3.5-Medium as the base model and train Vanilla SFT and all off-policy variants with LoRA under same optimization settings.

\paragraph{Baselines.}
For the on-policy setting, we report the pretrained SD3.5-Medium and compare \method\ against three groups of competitive baselines: (1) \emph{Single-task references}, including the task-specific Flow-GRPO models specialized for GenEval, OCR, and PickScore;
(2) \emph{Multi-objective RL}, represented by Flow-GRPO-Mix, which jointly optimizes the three target capabilities; and (3) \emph{KL-based OPD}, implemented as Vanilla OPD following the transition-kernel KL objective of Flow-OPD~\citep{fang2026flow}. For the off-policy setting, Vanilla SFT with single-state velocity supervision serves as the baseline.

\paragraph{Evaluation.}
Following the evaluation protocol of Flow-GRPO, we evaluate in-domain performance on GenEval, OCR, and PickScore. We further assess out-of-distribution compositional generalization on T2I-CompBench~\citep{huang2023t2i}, and evaluate general image quality and human preference on DrawBench~\citep{saharia2022photorealistic} using ImageReward, Aesthetic Score, UnifiedReward, and HPS-v2.1. Full training and evaluation details are provided in Appendices~\ref{app:training_details} and~\ref{app:evaluation}.

\subsection{Main Results}
\label{sec:main_performance}

Table~\ref{tab:main_results} reports the main results under the on-policy
and off-policy settings, we first examine multi-objective alignment in the
on-policy setting. The task-specific Flow-GRPO models perform strongly on their respective targets but exhibit substantial trade-offs on the others.
Flow-GRPO-Mix improves the balance across objectives, yet fails to match the specialized capabilities of individual experts. In contrast,
\method\ achieves balanced performance across all three objectives,
matching the GenEval and OCR experts while retaining competitive
PickScore performance. These results demonstrate the effectiveness of the trajectory-derived formulation for multi-reference OPD.

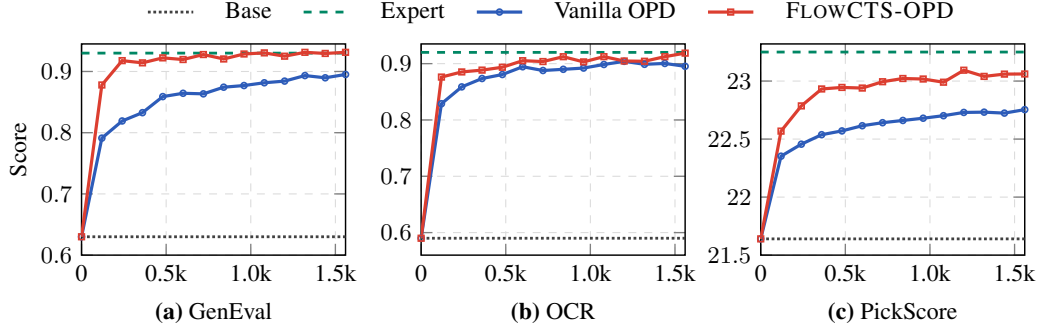
\begin{figure}[t]
    \centering
    \input{fig/training_curves.tex}
    \caption{
        Training dynamics of Vanilla OPD and \method\ across three task-specific
        metrics.
        The lower dotted horizontal lines denote the base-model performance,
        while the upper dashed horizontal lines denote the corresponding
        expert performance.
        Blue lines with circular markers represent OPD-KL, and red lines
        with square markers represent CTS.
    }
    \label{fig:training-curves}
\end{figure}

To isolate the effect of the trajectory-derived objective, we compare \method\ with Vanilla OPD within the same multi-reference on-policy pipeline. \method\ improves GenEval from $0.90$ to $0.93$, OCR from $0.90$ to $0.92$, and PickScore from $22.75$ to $23.06$. Figure~ \ref{fig:training-curves} further shows that \method\ converges faster and
reaches higher performance across all three metrics, approaching the corresponding GenEval and OCR expert levels substantially earlier than
Vanilla OPD.
Because the two methods differ only in their supervision
objectives, these controlled results suggest that the trajectory-derived velocity-matching objective is more effective than Vanilla OPD.

The off-policy results provide complementary evidence under a different starting-state distribution. Compared with Vanilla SFT, the FlowCTS-SFT objectives mainly improve OCR while maintaining comparable GenEval and PickScore performance, suggesting that the same trajectory-derived formulation also provides useful supervision in the off-policy setting. The effect of the supervision step in both settings is analyzed in
Section~\ref{sec:horizon}.

\begin{table}[t]
\caption{Main results for multi-reference on-policy distillation and its off-policy supervision extension. Higher values are better for all metrics. Bold and underlined values indicate the best and second-best results within each supervision setting, respectively.}
\label{tab:main_results}
\centering
\small
\setlength{\tabcolsep}{4.5pt}
\renewcommand{\arraystretch}{1.08}

\begin{tabular}{@{}lcccc@{}}
\toprule
\textbf{Method}
& \textbf{GenEval} $\uparrow$
& \textbf{OCR} $\uparrow$
& \textbf{PickScore} $\uparrow$
& \textbf{Overall} $\uparrow$ \\
\midrule

% ==================== On-policy setting ====================
\rowcolor{black!14}
\multicolumn{5}{@{}l}{
\textbf{\textit{I. On-policy multi-objective alignment}}
} \\

\addlinespace[2pt]
\rowcolor{black!5}
\multicolumn{5}{@{}l}{
\hspace{0.4em}\textit{Pretrained model}
} \\
\quad SD3.5-Medium
& 0.63
& 0.59
& 21.72
& 0.685 \\

\addlinespace[3pt]
\rowcolor{black!5}
\multicolumn{5}{@{}l}{
\hspace{0.4em}\textit{Reinforcement-learning baselines}
} \\
\quad FlowGRPO (GenEval)
& \underline{0.93}
& 0.65
& 21.60
& 0.804 \\
\quad FlowGRPO (OCR)
& 0.66
& \textbf{0.92}
& 21.74
& 0.805 \\
\quad FlowGRPO (PickScore)
& 0.51
& 0.70
& \textbf{23.25}
& 0.701 \\
\quad FlowGRPO (Mixed rewards)
& 0.73
& 0.83
& 21.84
& 0.800 \\

\addlinespace[3pt]
\rowcolor{black!5}
\multicolumn{5}{@{}l}{
\hspace{0.4em}\textit{KL-based on-policy distillation}
} \\
\quad Vanilla OPD w/ KL ($K=1$)
& 0.90
& \underline{0.90}
& 22.75
& 0.890 \\

\addlinespace[3pt]
\rowcolor{black!5}
\multicolumn{5}{@{}l}{
\hspace{0.4em}\textit{One-step velocity supervision}
} \\
\quad \textsc{FlowCTS-OPD} ($K=1$)
& \underline{0.93}
& \textbf{0.92}
& \underline{23.06}
& \underline{0.912} \\

\addlinespace[3pt]
\rowcolor{black!5}
\multicolumn{5}{@{}l}{
\hspace{0.4em}\textit{Multi-step trajectory supervision}
} \\
\quad FlowCTS-OPD ($K=2$)
& \textbf{0.94}
& \textbf{0.92}
& 22.96
& \textbf{0.914} \\
\quad FlowCTS-OPD ($K=3$)
& \underline{0.93}
& \textbf{0.92}
& 22.60
& 0.906 \\

% ==================== off-policy setting ====================
\addlinespace[3pt]
\midrule
\rowcolor{black!14}
\multicolumn{5}{@{}l}{
\textbf{\textit{II. Off-policy trajectory supervision}}
} \\

\addlinespace[2pt]
\rowcolor{black!5}
\multicolumn{5}{@{}l}{
\hspace{0.4em}\textit{Single-step supervised fine-tuning}
} \\
\quad Vanilla SFT
& \underline{0.71}
& 0.70
& \underline{21.68}
& 0.747 \\

\addlinespace[3pt]
\rowcolor{black!5}
\multicolumn{5}{@{}l}{
\hspace{0.4em}\textit{Multi-step trajectory supervision}
} \\
\quad FlowCTS-SFT ($K=2$)
& \textbf{0.72}
& 0.73
& 21.66
& \underline{0.761} \\
\quad FlowCTS-SFT ($K=3$)
& \underline{0.71}
& \textbf{0.75}
& \textbf{21.71}
& \textbf{0.763} \\
\quad FlowCTS-SFT ($K=4$)
& 0.70
& \underline{0.74}
& 21.63
& 0.756 \\

\bottomrule
\end{tabular}
\end{table}

\subsection{Effect of the Supervision step in OPD}
\label{sec:horizon}
We next study how far reference supervision within \method\ should extend beyond a student-visited state. As shown in Table~\ref{tab:main_results}, we compare different supervision step under both on-policy and off-policy settings. The step $K$ ranges from velocity matching at the current state when $K=1$ to supervision over a short student--reference trajectory segment when $K>1$.

Under the on-policy setting, extending the supervision horizon from $K=1$ to $K=2$ slightly improves GenEval from 0.93 to 0.94, yielding performance beyond that of the reference, while preserving the OCR score with only a minor reduction in PickScore. Extending the horizon further to $K=3$ provides no additional benefit and leads to a larger drop in PickScore. These results suggest that the supervision horizon should be chosen carefully: short-horizon objectives already capture the main training signal from the trajectory, while further extending the horizon leads to inconsistent improvements across metrics. To control for training duration, we further compare $K=1$ and $K=2$ under a doubled training-step budget. $K=1$ still underperforms $K=2$, suggesting that the improvement cannot be attributed solely to longer optimization. See Appendix~\ref{app:training_budget}.

The off-policy setting shows a similar pattern. Compared with Vanilla SFT, extending the supervision step to $K=2$ and $K=3$ improves all three metrics, with the most pronounced gains observed on OCR, which increases from 0.70 to 0.73 and 0.75, respectively. However, increasing it further to $K=4$ reduces all three metrics relative to $K=3$, indicating that the gains from extending the supervision horizon are not sustained at larger $K$.

We attribute this non-monotonic behavior to a trade-off between richer trajectory supervision and increased optimization difficulty. A larger $K$ provides reference signals at more future states, promoting structurally consistent evolution and improving compositional and spatial correctness. However, longer student--reference rollouts amplify the effect of early prediction errors, making later supervision harder to optimize. Stronger cross-step constraints may also suppress fine-grained details, weakening perceptual preference. Thus, the supervision horizon should be treated as an explicit design choice rather than simply maximized.

\subsection{Temporal Misallocation of KL Supervision}
\label{sec:kl_weighting}

We next analyze why the \method\ optimizes more effectively than Vanilla OPD. In Vanilla OPD, the deterministic ODE dynamics are
first reformulated as stochastic SDE transitions, and the KL divergence between the resulting transition kernels reduces to velocity regression with a time-dependent weight, where $w_{\mathrm{KL}}(t)$ is induced by the auxiliary SDE transition variance.
\begin{equation}
D_{\mathrm{KL}}
\left(
\pi_{\theta}(\cdot \mid x_t)
\| 
\pi_{\mathrm{ref}}(\cdot \mid x_t)
\right)
=
w_{\mathrm{KL}}(t)
\left\|
v_{\theta}(x_t,t,c)-v_{\mathrm{ref}}(x_t,t,c)
\right\|_2^2,
\end{equation}

\begin{figure}[t]
    \centering
    \input{fig/trajectory_weight_diagnostics.tex}
    \caption{
    Temporal signal allocation in KL-based OPD. Both horizontal axes denote
    normalized denoising progress from noise to image. The two quantities are
    computed from the same student-visited states and training observations.
    (a) Effective per-step KL loss: $w_{\mathrm{KL}}(t) \lVert v_{\theta}(x_t,t,c)-v_{\mathrm{ref}}(x_t,t,c)\rVert_2^2$. (b) Mean squared velocity gap without the additional KL-induced weight: $\lVert v_{\theta}(x_t,t,c)-v_{\mathrm{ref}}(x_t,t,c)\rVert_2^2$. The KL objective concentrates its effective supervision near the image endpoint, while the underlying student--reference velocity mismatch is largest during earlier denoising stages.
    }
    \label{fig:trajectory-weight-diagnostics}
\end{figure}
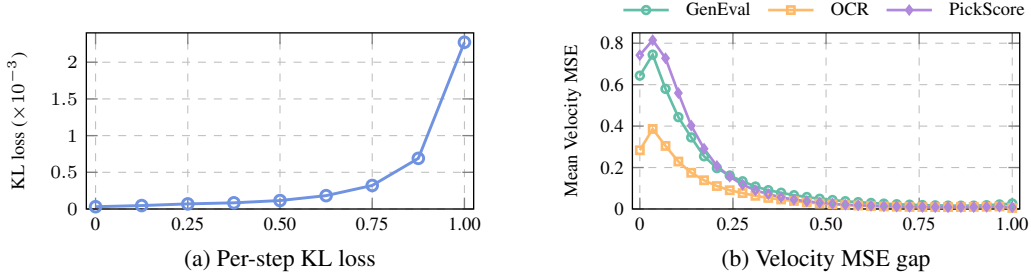

Figure~\ref{fig:trajectory-weight-diagnostics} reveals a clear mismatch in temporal signal allocation. The KL loss is strongly concentrated near the image endpoint, whereas the largest velocity-field differences between the pretrained model and the three reference experts consistently occur during early denoising. This observation is consistent with TempFlow-GRPO~\citep{he2025tempflow}, which shows that perturbations during early structural stages have a substantially greater influence
on the final generation outcome than those during late refinement
stages.

Therefore, the SDE-induced KL weight emphasizes late timesteps while underweighting the stages where the reference models introduce their largest corrections. In contrast, our $K=1$ objective directly matches velocities at student-visited states without this temporal weighting. This difference provides an explanation for its faster convergence and stronger final performance in Figure~\ref{fig:training-curves}.

\subsection{Capability and Generalization Analysis}
\label{sec:capability}

We further examine whether the \method\ objective generalizes
beyond the three training metrics. As shown in Table~\ref{tab:general_quality}, \method\ consistently outperforms Vanilla OPD across ImageReward, Aesthetic Score, UnifiedReward, and HPS-v2.1. These results show that its improvements are not limited to the training targets, but extend to broader measures of visual quality and human preference.

\begin{table}[t]
\centering
\caption{\textbf{Results on T2I-CompBench.}
The best and second-best results within the OPD group are shown in
bold and underlined, respectively.}
\label{tab:ood_robustness}

\small
\setlength{\tabcolsep}{3.5pt}
\renewcommand{\arraystretch}{0.98}

\begin{tabular*}{\linewidth}{
    @{\extracolsep{\fill}}
    lccccccc
    @{}
}
\toprule
Model & Color & Shape & Texture & Complex & 3D-Spatial & Numeracy & Non-Spatial \\
\midrule

\multicolumn{8}{l}{\textit{Pretrained model}} \\
\quad SD3.5-M
& 0.7994
& 0.5669
& 0.7338
& 0.3800
& 0.3739
& 0.5927
& 0.3146 \\

\midrule
\multicolumn{8}{l}{\textit{Reinforcement learning}} \\
\quad GRPO-Mix
& 0.7966
& 0.5803
& 0.7392
& 0.3677
& 0.3681
& 0.6388
& 0.3130 \\

\midrule
\multicolumn{8}{l}{\textit{On-policy distillation}} \\
\quad Vanilla OPD
& \textbf{0.8350}
& \underline{0.6211}
& \textbf{0.7508}
& \underline{0.3834}
& \underline{0.4411}
& \underline{0.6840}
& \textbf{0.3135} \\

\quad \method
& \underline{0.8295}
& \textbf{0.6285}
& \underline{0.7414}
& \textbf{0.3839}
& \textbf{0.4433}
& \textbf{0.6909}
& \underline{0.3096} \\

\bottomrule
\end{tabular*}
\end{table}

\begin{table}[t]
\centering
\caption{Comparison on general image quality and human preference metrics.The best and second-best results within the OPD group are shown in bold and underlined, respectively.
\textsuperscript{\ddag} Evaluated at a resolution of $1024\times1024$.}
\label{tab:general_quality}

\small
\setlength{\tabcolsep}{9pt}
\renewcommand{\arraystretch}{1.0}

\begin{tabular}{@{}lcccc@{}}
\toprule
Model & ImageReward & Aesthetic & UnifiedReward & HPS-v2.1 \\
\midrule

\multicolumn{5}{@{}l}{\textit{Pretrained models}} \\
\quad SD-XL\textsuperscript{\ddag}
& 0.76 & 5.60 & 2.93 & 0.28 \\
\quad SD3.5-L\textsuperscript{\ddag}
& 0.96 & 5.50 & 3.25 & 0.29 \\
\quad FLUX.1-Dev
& 0.96 & 5.71 & 3.27 & 0.27 \\
\quad SD3.5-M
& 0.82 & 5.37 & 3.00 & 0.28 \\

\midrule
\multicolumn{5}{@{}l}{\textit{On-policy distillation}} \\
\quad Vanilla OPD
& \underline{1.21}
& \underline{5.46}
& \underline{3.28}
& \underline{0.30} \\
\quad \method
& \textbf{1.27}
& \textbf{5.64}
& \textbf{3.32}
& \textbf{0.31} \\

\bottomrule
\end{tabular}
\end{table}

Table~\ref{tab:ood_robustness} further evaluates out-of-distribution compositional capabilities on T2I-CompBench. Compared with Vanilla OPD, \method\ performs better on shape, spatial relations, and numeracy, while the differences on color and texture remain limited. This pattern is consistent with the temporal analysis in Section~\ref{sec:kl_weighting}: since early denoising stages play a greater role in establishing global structure, correcting the under-allocation of supervision at these stages is expected to benefit structural and relational capabilities more than visual attributes. These results suggest that the allocation of supervision across timesteps plays an important role in effective post-training.

\section{Related Work}

\paragraph{Post-training for Flow Models.}
Post-training has become an important approach for adapting pretrained visual generative models toward downstream objectives. Early methods such as DDPO and DPOK optimize diffusion models through policy-gradient updates using terminal rewards, ReFL directly backpropagates differentiable reward feedback, while Diffusion-DPO performs alignment from offline preference pairs ~\citep{black2024training,fan2023dpok,xu2023imagereward,wallace2024diffusion,clark2024directly,liu2026improving}. Building on these developments, more recent work introduces GRPO-style online optimization for modern diffusion and rectified-flow generators. DanceGRPO~\citep{xue2025dancegrpo} develops a unified GRPO framework across diffusion and rectified-flow generators,
while Flow-GRPO enables online policy-gradient optimization of flow models through stochastic sampling trajectories ~\citep{liu2026flow}. TempFlow-GRPO~\citep{he2025tempflow} further introduces temporally aware credit assignment for flow-model optimization, whereas DiffusionNFT ~\citep{zheng2025diffusionnft} incorporates online reward feedback into forward-process flow matching. 
Complementary to these reward and preference-driven approaches, our work studies reference-guided on-policy distillation and focuses on how dense supervision should be constructed for flow models.

\paragraph{On-Policy Distillation for Flow Models.}
On-policy distillation was initially developed for autoregressive
language models, where the student generates its own sequences and receives dense reference supervision at the visited prefixes.
MiniLLM optimizes a reverse-KL objective on student rollouts, while GKD generalizes this paradigm to different divergence objectives and sampling strategies~\citep{gu2024minillm,agarwal2024policy}.Recent studies further use OPD to consolidate capabilities from multiple specialized references into a unified student
~\citep{yang2026learning,ma2026mopd,zeng2026glm,yang2025qwen3,xiao2026mimo,li2026kat}. Recent works extend this idea to flow models. Flow-OPD and DiffusionOPD construct per-step KL objectives on student-generated states~\citep{fang2026flow, li2026diffusionopd}. Concurrent to our work, DanceOPD independently adopts direct velocity matching at a student-visited state and motivates it as a local field-regression objective~\citep{zhou2026danceopd}. FlowCTS instead derives a family of objectives from continuous student--reference trajectory differences. It further analyzes the temporal misallocation of transition-kernel KL relative to reference-student velocity errors, studies different supervision step, and extends the same formulation to off-policy supervision.

\paragraph{Trajectory Distillation for Flow Models.}
Most diffusion distillation methods focus on step reduction, compressing a multi-step reference into a few-step student for efficient inference ~\citep{salimans2022progressive,liu2023instaflow}.
Trajectory-based methods imitate intermediate transitions or enforce consistency across timesteps ~\citep{song2023consistency,kim2024consistency,boffi2024flow},
while DMD methods align the student and reference
distributions without requiring trajectory-wise correspondence
~\citep{yin2024improved,yin2024one}. In contrast, \method\ does not learn a shortcut generator. It uses finite trajectory segments as supervision units for post-training the original time-dependent velocity field and studies how the supervision horizon affects the resulting model.

\section{Conclusion}
\label{sec:conclusion}

We study on-policy distillation for flow matching models and derive
\method\ from continuous student and reference trajectories initialized at shared student-visited states. The resulting objectives cover both local velocity matching and short-horizon supervision. In a controlled multi-reference setting, \method\ converges faster and outperforms transition-kernel KL. Longer horizons yield non-monotonic gains, reflecting a trade-off between additional trajectory information and optimization difficulty. Temporal analysis further reveals that KL supervision is concentrated at late denoising stages, despite larger reference-field corrections occurring earlier. Together, these results identify temporal weighting and supervision horizon as two key design choices for OPD in continuous-time generative models.

\bibliography{iclr2026_conference}
\bibliographystyle{iclr2026_conference}

\appendix
\section{Derivation of the Velocity Upper Bound}
\label{app:upper_bound}

We provide the derivation of the velocity upper bound used in Section~\ref{sec:method}. 
Consider a reverse-time segment $[t_{\mathrm{e}},t_{\mathrm{s}}]$ with 
$0 \leq t_{\mathrm{e}} < t_{\mathrm{s}} \leq 1$. 
The student and reference trajectories start from the same state $\vx_{t_{\mathrm{s}}}$ and are defined as
\begin{equation}
\vx_t^{\vtheta}
=
\vx_{t_{\mathrm{s}}}
-
\int_t^{t_{\mathrm{s}}}
\vv_{\vtheta}(\vx_r^{\vtheta},r,c)\,dr,
\qquad
\vx_t^{\mathrm{ref}}
=
\vx_{t_{\mathrm{s}}}
-
\int_t^{t_{\mathrm{s}}}
\vv_{\mathrm{ref}}(\vx_r^{\mathrm{ref}},r,c)\,dr .
\end{equation}
Subtracting the two trajectories gives
\begin{equation}
\vx_t^{\vtheta}
-
\vx_t^{\mathrm{ref}}
=
-
\int_t^{t_{\mathrm{s}}}
\left[
\vv_{\vtheta}(\vx_r^{\vtheta},r,c)
-
\vv_{\mathrm{ref}}(\vx_r^{\mathrm{ref}},r,c)
\right]dr .
\end{equation}
For simplicity, define the velocity mismatch along the two trajectories as
\begin{equation}
\Delta \vv_r
=
\vv_{\vtheta}(\vx_r^{\vtheta},r,c)
-
\vv_{\mathrm{ref}}(\vx_r^{\mathrm{ref}},r,c).
\end{equation}
Then
\begin{equation}
\vx_t^{\vtheta}
-
\vx_t^{\mathrm{ref}}
=
-
\int_t^{t_{\mathrm{s}}}
\Delta \vv_r\,dr .
\end{equation}

The continuous trajectory objective is
\begin{equation}
\mathcal{L}_{\mathrm{CTS}}(\vtheta)
=
\E_{\vx_{t_{\mathrm{s}}},c}
\left[
\int_{t_{\mathrm{e}}}^{t_{\mathrm{s}}}
\left\lVert
\vx_t^{\vtheta}
-
\vx_t^{\mathrm{ref}}
\right\rVert^2
dt
\right].
\end{equation}
% Substituting the integral form of the state discrepancy gives
% \begin{equation}
% \mathcal{L}_{\mathrm{CTS}}(\vtheta)
% =
% \E_{\vx_{t_{\mathrm{s}}},c}
% \left[
% \int_{t_{\mathrm{e}}}^{t_{\mathrm{s}}}
% \left\lVert
% \int_t^{t_{\mathrm{s}}}
% \Delta \vv_r\,dr
% \right\rVert^2
% dt
% \right].
% \end{equation}
% By Cauchy--Schwarz,
% \begin{equation}
% \left\lVert
% \int_t^{t_{\mathrm{s}}}
% \Delta \vv_r\,dr
% \right\rVert^2
% \leq
% (t_{\mathrm{s}}-t)
% \int_t^{t_{\mathrm{s}}}
% \left\lVert
% \Delta \vv_r
% \right\rVert^2
% dr .
% \end{equation}
% Therefore,
% \begin{equation}
% \mathcal{L}_{\mathrm{CTS}}(\vtheta)
% \leq
% \E_{\vx_{t_{\mathrm{s}}},c}
% \left[
% \int_{t_{\mathrm{e}}}^{t_{\mathrm{s}}}
% (t_{\mathrm{s}}-t)
% \int_t^{t_{\mathrm{s}}}
% \left\lVert
% \Delta \vv_r
% \right\rVert^2
% drdt
% \right].
% \end{equation}
% The integration region is
% \begin{equation}
% t_{\mathrm{e}} \leq t \leq r \leq t_{\mathrm{s}}.
% \end{equation}
% Exchanging the order of integration yields
% \begin{equation}
% \mathcal{L}_{\mathrm{CTS}}(\vtheta)
% \leq
% \E_{\vx_{t_{\mathrm{s}}},c}
% \left[
% \int_{t_{\mathrm{e}}}^{t_{\mathrm{s}}}
% \left(
% \int_{t_{\mathrm{e}}}^{r}
% (t_{\mathrm{s}}-t)dt
% \right)
% \left\lVert
% \Delta \vv_r
% \right\rVert^2
% dr
% \right].
% \end{equation}

Substituting the integral form of the state discrepancy gives
\begin{equation}
\begin{aligned}
\mathcal{L}_{\mathrm{CTS}}(\vtheta)
&=
\E_{\vx_{t_{\mathrm{s}}},c}
\left[
\int_{t_{\mathrm{e}}}^{t_{\mathrm{s}}}
\left\lVert
\int_t^{t_{\mathrm{s}}}
\Delta \vv_r\,dr
\right\rVert^2
dt
\right]
\\
&\leq
\E_{\vx_{t_{\mathrm{s}}},c}
\left[
\int_{t_{\mathrm{e}}}^{t_{\mathrm{s}}}
(t_{\mathrm{s}}-t)
\int_t^{t_{\mathrm{s}}}
\left\lVert
\Delta \vv_r
\right\rVert^2
dr\,dt
\right]
\\
&=
\E_{\vx_{t_{\mathrm{s}}},c}
\left[
\int_{t_{\mathrm{e}}}^{t_{\mathrm{s}}}
\left(
\int_{t_{\mathrm{e}}}^{r}
(t_{\mathrm{s}}-t)\,dt
\right)
\left\lVert
\Delta \vv_r
\right\rVert^2
dr
\right].
\end{aligned}
\end{equation}
The inner trajectory integral is first bounded by Cauchy--Schwarz, after which the resulting double integral is reordered. The trajectory-derived weight is
\begin{equation}
w(r)
=
\int_{t_{\mathrm{e}}}^{r}
(t_{\mathrm{s}}-t)dt
=
\frac{
(t_{\mathrm{s}}-t_{\mathrm{e}})^2
-
(t_{\mathrm{s}}-r)^2
}{2}.
\end{equation}
Substituting $\Delta \vv_r$ back gives the velocity upper-bound objective
\begin{equation}
\mathcal{L}_{\mathrm{CTS}}(\vtheta)
\leq
\mathcal{L}_{\mathrm{VUB}}(\vtheta)
=
\E_{\vx_{t_{\mathrm{s}}},c}
\left[
\int_{t_{\mathrm{e}}}^{t_{\mathrm{s}}}
w(r)
\left\lVert
\vv_{\vtheta}(\vx_r^{\vtheta},r,c)
-
\vv_{\mathrm{ref}}(\vx_r^{\mathrm{ref}},r,c)
\right\rVert^2
dr
\right].
\end{equation}

\section{Derivation of the Discrete Training Objective}
\label{app:discrete_objective}

We derive the discrete form of the trajectory objective. 
We discretize the reverse-time segment $[t_{\mathrm{e}},t_{\mathrm{s}}]$ into $K$ uniform Euler steps. 
Let
\begin{equation}
h=\frac{t_{\mathrm{s}}-t_{\mathrm{e}}}{K},
\qquad
t_i=t_{\mathrm{s}}-ih,
\qquad
i=0,\dots,K.
\end{equation}
The student and reference rollouts start from the same state:
\begin{equation}
\vx_0^{\vtheta}
=
\vx_0^{\mathrm{ref}}
=
\vx_{t_{\mathrm{s}}}.
\end{equation}
Using Euler discretization, the two rollouts are
\begin{equation}
\vx_{i+1}^{\vtheta}
=
\vx_i^{\vtheta}
-
h\vv_{\vtheta}(\vx_i^{\vtheta},t_i,c),
\qquad
\vx_{i+1}^{\mathrm{ref}}
=
\vx_i^{\mathrm{ref}}
-
h\vv_{\mathrm{ref}}(\vx_i^{\mathrm{ref}},t_i,c),
\end{equation}
for $i=0,\dots,K-1$.

Define the velocity mismatch at step $i$ as
\begin{equation}
\delta_i
=
\vv_{\vtheta}(\vx_i^{\vtheta},t_i,c)
-
\vv_{\mathrm{ref}}(\vx_i^{\mathrm{ref}},t_i,c).
\end{equation}
Then the state discrepancy follows
\begin{equation}
\vx_{i+1}^{\vtheta}
-
\vx_{i+1}^{\mathrm{ref}}
=
\vx_i^{\vtheta}
-
\vx_i^{\mathrm{ref}}
-
h\delta_i .
\end{equation}
Since $\vx_0^{\vtheta}-\vx_0^{\mathrm{ref}}=\vzero$, we have
\begin{equation}
\vx_i^{\vtheta}
-
\vx_i^{\mathrm{ref}}
=
-
h
\sum_{j=0}^{i-1}
\delta_j,
\qquad
i=1,\dots,K.
\end{equation}

A discrete state-space trajectory loss can be written as
\begin{equation}
\mathcal{L}_{\mathrm{state,disc}}
=
h
\sum_{i=1}^{K}
\left\lVert
\vx_i^{\vtheta}
-
\vx_i^{\mathrm{ref}}
\right\rVert^2 .
\end{equation}
Substituting the accumulated error form gives
\begin{equation}
\mathcal{L}_{\mathrm{state,disc}}
=
h^3
\sum_{i=1}^{K}
\left\lVert
\sum_{j=0}^{i-1}
\delta_j
\right\rVert^2 .
\end{equation}
By the discrete Cauchy--Schwarz inequality,
\begin{equation}
\left\lVert
\sum_{j=0}^{i-1}
\delta_j
\right\rVert^2
\leq
i
\sum_{j=0}^{i-1}
\left\lVert
\delta_j
\right\rVert^2 .
\end{equation}
Therefore,
\begin{equation}
\mathcal{L}_{\mathrm{state,disc}}
\leq
h^3
\sum_{i=1}^{K}
i
\sum_{j=0}^{i-1}
\left\lVert
\delta_j
\right\rVert^2 .
\end{equation}
Exchanging the summation order, we obtain
\begin{equation}
\mathcal{L}_{\mathrm{state,disc}}
\leq
h^3
\sum_{j=0}^{K-1}
\left(
\sum_{i=j+1}^{K}
i
\right)
\left\lVert
\delta_j
\right\rVert^2 .
\end{equation}
The inner sum has the closed form
\begin{equation}
\sum_{i=j+1}^{K}
i
=
\frac{K(K+1)-j(j+1)}{2}.
\end{equation}
Thus,
\begin{equation}
\mathcal{L}_{\mathrm{state,disc}}
\leq
h^3
\sum_{j=0}^{K-1}
\frac{K(K+1)-j(j+1)}{2}
\left\lVert
\delta_j
\right\rVert^2 .
\end{equation}

The discrete upper bound determines the relative contribution of the velocity mismatch at each supervised state. For a fixed supervision horizon $K$ and trajectory segment, $h^3$ is a positive constant independent of $\vtheta$. It therefore does not affect the minimizer or the relative weighting across timesteps, and can be absorbed into the overall loss coefficient. We define
the unnormalized trajectory-derived weights as
\begin{equation}
\alpha_j
=
\frac{K(K+1)-j(j+1)}{2},
\qquad
j=0,\dots,K-1.
\label{eq:discrete_weight}
\end{equation}
For training, we normalize these coefficients as
\begin{equation}
\bar{\alpha}_j
=
\frac{\alpha_j}
{\sum_{i=0}^{K-1}\alpha_i},
\qquad
\sum_{i=0}^{K-1}\alpha_i
=
\frac{K(K+1)(2K+1)}{6},
\label{eq:normalized_discrete_weight}
\end{equation}
which preserves the trajectory-derived temporal profile while keeping the overall loss scale comparable across different values of $K$. The practical objective is therefore
\begin{equation}
\mathcal{L}_{\mathrm{CTS}}^{(K)}(\vtheta)
=
\E
\left[
\sum_{j=0}^{K-1}
\bar{\alpha}_j
\left\lVert
\vv_{\vtheta}(\vx_j^{\vtheta},t_j,c)
-
\vv_{\mathrm{ref}}(\vx_j^{\mathrm{ref}},t_j,c)
\right\rVert_2^2
\right].
\label{eq:practical_discrete_cts}
\end{equation}
When $K=1$, $\bar{\alpha}_0=1$, and the objective reduces to velocity
matching at the segment starting state.

\section{Training Details}
\label{app:training_details}

\subsection{On-Policy multi-reference Distillation}

\paragraph{Implementation details.}
We use Stable Diffusion 3.5 Medium as the base model and initialize the student from the averaged multi-task Flow-GRPO LoRA checkpoint. All models are fine-tuned with LoRA applied to the attention projections of the SD3 transformer, including the query, key, value, output, and added-conditioning projections. We set the LoRA rank to 32 and LoRA alpha to 64. Training is performed on four H200 GPUs.

Student rollouts are generated at a resolution of $512\times512$ using 10 sampling steps and a classifier-free guidance scale of 4.5. A timestep fraction of 0.99 is used, resulting in nine denoising transitions for optimization. The sampling batch size is 16 per GPU, with four sampling batches per epoch and gradient accumulation over two steps. We optimize the student with AdamW using a learning rate of $1\times10^{-4}$, weight decay
$1\times10^{-4}$, and a maximum gradient norm of 1.0. We maintain an exponential moving average of the trainable parameters with decay 0.9, updated every eight optimization steps.

\paragraph{Prompt routing and reference models.}
We use the OCR, GenEval, and PickScore prompt sets released with Flow-GRPO. Each mini-batch is drawn from one prompt set and routed to its corresponding task-specific reference model: the text-rendering, GenEval, or PickScore Flow-GRPO expert. The three datasets are sampled alternately with a cycle ratio of $1{:}3{:}1$ for OCR, GenEval, and PickScore, respectively. No additional auxiliary is introduced.

\paragraph{OPD objectives.}
Vanilla OPD and \method use the same initialization, prompt mixture, student rollouts, reference routing, batch construction, and optimization hyperparameters. Vanilla OPD uses the step-wise transition-kernel KL as the distillation objective.

For \method, we replace the transition-kernel policy objective with the trajectory-derived velocity objective in Eq.~\ref{eq:discrete_cts}. When $K=1$, the objective reduces to mean-squared velocity matching between the student and the routed reference model at a student-visited state. For $K>1$, the student and reference models are rolled out over the subsequent trajectory segment, and the per-step losses are weighted according to
Eq.~\ref{eq:discrete_weight}.

\subsection{off-policy Extension}

For the off-policy experiments, we construct training trajectories from 50K image--text pairs sampled from Fine-T2I and use SD3.5-Medium as the base model. Vanilla SFT and all \methodoff variants are trained with LoRA under the same data and optimization configuration. Vanilla SFT corresponds to single-state velocity supervision, while \methodoff extends the objective to
$K\in\{2,3,4\}$ subsequent states using the trajectory-derived weights. All off-policy experiments are conducted on four A800 GPUs.

\subsection{Effect of SDE and ODE Sampling}
\label{app:sde_ode}

We compare SDE and ODE sampling for constructing the student-visited states during on-policy distillation. The two variants use the same velocity objective and training configuration, differing only in the rollout process used to obtain on-policy states.

\begin{figure}[t]
    \centering
    \input{fig/sde_ode.tex}
    \caption{
    Comparison of SDE and ODE sampling for constructing student-visited states during on-policy distillation. Both variants use the same velocity-matching objective and training configuration, differing only in the rollout process used to generate on-policy states. SDE sampling consistently achieves higher GenEval performance across training checkpoints, with a larger advantage during early optimization. The horizontal dashed lines denote the pretrained base model and the     corresponding task-specific reference.
    }
    \label{fig:sde_ode}
\end{figure}
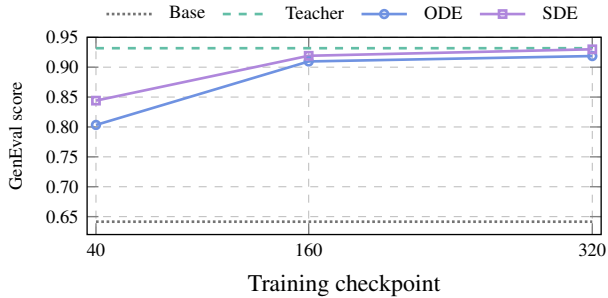

As shown in Figure~\ref{fig:sde_ode}, SDE sampling consistently
outperforms deterministic ODE sampling throughout training. The
difference is most pronounced at the early checkpoint, where SDE
improves GenEval from 0.80 to 0.84. The advantage remains at the end of training, with SDE reaching 0.93 compared with 0.92 for ODE sampling and approaching the corresponding reference performance. These results suggest that stochastic exploration improves the states encountered during on-policy distillation, particularly during early optimization.

\section{Evaluation}
\label{app:evaluation}

Unless otherwise specified, we generate images at a resolution of
$512\times512$ using 40 inference steps, a classifier-free guidance scale of 4.5, and a random seed of 42. All compared methods use identical prompts, generation settings, and evaluator implementations. Models marked with $\ddagger$ in Table~\ref{tab:general_quality} are evaluated at $1024\times1024$.

Following Flow-GRPO~\citep{liu2026flow}, we evaluate GenEval, OCR, and PickScore on their corresponding test sets. Compositional generalization is evaluated on the validation split of T2I-CompBench~\citep{huang2023t2i}, with ten images generated per prompt using the official evaluation implementation. General image quality and preference alignment are evaluated on DrawBench using ImageReward, Aesthetic Score, UnifiedReward, and HPS-v2.1,
following the Flow-GRPO evaluation protocol. All compared methods use identical generation and evaluator settings.

\section{Controlling for the Training Budget}
\label{app:training_budget}

Since \method~($K=2$) evaluates two supervised states per optimization step, we conduct a small-scale auxiliary study to efficiently control for the supervision budget. Specifically, we compare $K=2$ trained for $N$ steps with $K=1$ trained for $N$ and $2N$ steps, where the latter matches the total number of supervised state evaluations. All runs in this study use the same
reduced training configuration and are intended for controlled comparison within Table~\ref{tab:training_budget}.

\begin{table}[t]
\centering
\caption{
Training-budget control. The extended $K=1$ run matches the number of supervised state evaluations used by $K=2$.
}
\label{tab:training_budget}
\small
\setlength{\tabcolsep}{6pt}
\begin{tabular}{lcccc}
\toprule
Method & Training steps & GenEval $\uparrow$ & OCR $\uparrow$
& PickScore $\uparrow$ \\
\midrule
\method~($K=1$) & $N$  & 0.91 & 0.89 & 22.93 \\
\method~($K=1$) & $2N$ & 0.92 & 0.91 & 23.01 \\
\method~($K=2$) & $N$  & 0.93 & 0.91 & 22.80 \\
\bottomrule
\end{tabular}
\end{table}

As shown in Table~\ref{tab:training_budget}, doubling the number of training steps for $K=1$ produces modest improvements across all three metrics, but does not reproduce the behavior of $K=2$. The latter achieves a higher GenEval score and comparable OCR performance, while yielding a lower PickScore. This distinct metric profile indicates that the effect of $K=2$ cannot be explained solely by the number of supervised state evaluations. Instead, supervision at a subsequent rollout state places additional constraints on how the sample evolves along the trajectory, which particularly benefits compositional and structural correctness. Meanwhile, qualitative inspection suggests that this stronger trajectory constraint may suppress some fine-grained details, resulting in slightly smoother images and a lower preference score. We therefore view the result as a trade-off introduced by the supervision span, rather than a uniform improvement obtained from a larger training budget.

\end{document}

%% file: math_commands.tex
\usepackage{amsmath,amsfonts,bm}

\def\eqref#1{equation~\ref{#1}}
\def\1{\bm{1}}

\def\vzero{{\bm{0}}}

\def\vtheta{{\bm{\theta}}}

\def\vu{{\bm{u}}}
\def\vv{{\bm{v}}}

\def\vx{{\bm{x}}}

\def\mI{{\bm{I}}}

\DeclareMathAlphabet{\mathsfit}{\encodingdefault}{\sfdefault}{m}{sl}
\SetMathAlphabet{\mathsfit}{bold}{\encodingdefault}{\sfdefault}{bx}{n}

\newcommand{\E}{\mathbb{E}}

\newcommand{\KL}{D_{\mathrm{KL}}}

%% file: fig/training_curves.tex
\begin{tikzpicture}

% =========================================================
% (a) GenEval
% =========================================================
\begin{axis}[
    name=plot1,
    iclrfigure,
    width=0.25\linewidth,
    height=0.2\linewidth,
    scale only axis,
    grid=major,
    grid style={dashed,black!14},
    xmin=0,
    xmax=1560,
    ymin=0.60,
    ymax=0.945,
    xtick={0,500,1000,1500},
    xticklabels={0,0.5k,1.0k,1.5k},
    ytick={0.60,0.70,0.80,0.90},
    xlabel={\textbf{(a)} GenEval},
    xlabel style={yshift=-0.1em},
    ylabel={Score},
    ylabel style={yshift=-0.15em},
    restrict x to domain=0:1560,
    unbounded coords=discard
]

% Base: lower horizontal reference
\addplot[
    baseline,
    forget plot
] coordinates {
    (0,0.63)
    (1560,0.63)
};

% Expert: upper horizontal reference
\addplot[
    expertline,
    forget plot
] coordinates {
    (0,0.93)
    (1560,0.93)
};

% OPD-KL: blue circles
\addplot[
    opdcurve,
    forget plot
] table[
    x index=0,
    y index=1,
    col sep=space
] {fig/data/geneval_training.txt};

% CTS: red squares
\addplot[
    ctscurve,
    forget plot
] table[
    x index=0,
    y index=2,
    col sep=space
] {fig/data/geneval_training.txt};

\end{axis}

% =========================================================
% (b) OCR
% =========================================================
\begin{axis}[
    name=plot2,
    at={(plot1.north east)},
    anchor=north west,
    xshift=1.0cm,
    iclrfigure,
    width=0.25\linewidth,
    height=0.2\linewidth,
    scale only axis,
    grid=major,
    grid style={dashed,black!14},
    xmin=0,
    xmax=1560,
    ymin=0.56,
    ymax=0.935,
    xtick={0,500,1000,1500},
    xticklabels={0,0.5k,1.0k,1.5k},
    ytick={0.60,0.70,0.80,0.90},
    xlabel={\textbf{(b)} OCR},
    xlabel style={yshift=-0.1em},
    restrict x to domain=0:1560,
    unbounded coords=discard,
    legend columns=4,
    legend style={
        at={(0.50,1.05)},
        anchor=south,
        fill=none,
        draw=none,
        font=\footnotesize,
        column sep=0.95em
    }
]

% Base: lower horizontal reference
\addplot[
    baseline,
    forget plot
] coordinates {
    (0,0.59)
    (1560,0.59)
};

% Expert: upper horizontal reference
\addplot[
    expertline,
    forget plot
] coordinates {
    (0,0.92)
    (1560,0.92)
};

% OPD-KL: blue circles
\addplot[
    opdcurve,
    forget plot
] table[
    x index=0,
    y index=1,
    col sep=space
] {fig/data/ocr_training.txt};

% CTS: red squares
\addplot[
    ctscurve,
    forget plot
] table[
    x index=0,
    y index=2,
    col sep=space
] {fig/data/ocr_training.txt};

% Shared legend
\addlegendimage{baseline}
\addlegendentry{Base}

\addlegendimage{expertline}
\addlegendentry{Expert}

\addlegendimage{opdcurve}
\addlegendentry{Vanilla OPD}

\addlegendimage{ctscurve}
\addlegendentry{\textsc{FlowCTS-OPD}}

\end{axis}

% =========================================================
% (c) PickScore
% =========================================================
\begin{axis}[
    name=plot3,
    at={(plot2.north east)},
    anchor=north west,
    xshift=1.0cm,
    iclrfigure,
    width=0.25\linewidth,
    height=0.2\linewidth,
    scale only axis,
    grid=major,
    grid style={dashed,black!14},
    xmin=0,
    xmax=1560,
    ymin=21.50,
    ymax=23.32,
    xtick={0,500,1000,1500},
    xticklabels={0,0.5k,1.0k,1.5k},
    ytick={21.5,22.0,22.5,23.0},
    xlabel={\textbf{(c)} PickScore},
    xlabel style={yshift=-0.10em},
    restrict x to domain=0:1560,
    unbounded coords=discard
]

% Base: lower horizontal reference
\addplot[
    baseline,
    forget plot
] coordinates {
    (0,21.64)
    (1560,21.64)
};

% Expert: upper horizontal reference
\addplot[
    expertline,
    forget plot
] coordinates {
    (0,23.25)
    (1560,23.25)
};

% OPD-KL: blue circles
\addplot[
    opdcurve,
    forget plot
] table[
    x index=0,
    y index=1,
    col sep=space
] {fig/data/pickscore_training.txt};

% CTS: red squares
\addplot[
    ctscurve,
    forget plot
] table[
    x index=0,
    y index=2,
    col sep=space
] {fig/data/pickscore_training.txt};

\end{axis}

\end{tikzpicture}

%% file: fig/trajectory_weight_diagnostics.tex
\begin{tikzpicture}
{\footnotesize

% =========================================================
% (a) Per-step KL loss
% =========================================================
\begin{axis}[
    name=plot1,
    width=0.48\linewidth,
    height=0.28\linewidth,
    grid=major,
    grid style={dashed,black!25},
    xmin=-0.2,
    xmax=8.2,
    ymin=0,
    ymax=2.4,
    xtick={0,2,4,6,8},
    xticklabels={
        \scriptsize 0,
        \scriptsize 0.25,
        \scriptsize 0.50,
        \scriptsize 0.75,
        \scriptsize 1.00
    },
    ytick={0,0.5,1.0,1.5,2.0},
    tick label style={font=\scriptsize},
    xlabel={\footnotesize (a) Per-step KL loss},
    xlabel style={yshift=-0.1em},
    ylabel={KL loss ($\times 10^{-3}$)},
    ylabel style={
        font=\scriptsize,
        at={(axis description cs:-0.13,0.5)},
        anchor=south
    },
    scaled ticks=false,
    axis line style={line width=0.5pt},
    tick style={line width=0.5pt}
]

\addplot[
    color5,
    mark=o,
    mark size=2.0pt,
    line width=1.1pt,
    mark options={
        solid,
        fill=color5,
        draw=color5
    }
] table[
    x index=0,
    y expr=1000*\thisrowno{1},
    col sep=space
] {fig/data/kl_loss_ckpt40.txt};

\end{axis}

% =========================================================
% (b) Velocity cosine gap
% =========================================================
\begin{axis}[
    name=plot2,
    at={(plot1.north east)},
    anchor=north west,
    xshift=2.1cm,
    width=0.48\linewidth,
    height=0.28\linewidth,
    grid=major,
    grid style={dashed,black!25},
    xmin=-0.02,
    xmax=1.02,
    ymin=0,
    ymax=0.85,
    xtick={0,0.25,0.50,0.75,1.00},
    xticklabels={
        \scriptsize 0,
        \scriptsize 0.25,
        \scriptsize 0.50,
        \scriptsize 0.75,
        \scriptsize 1.00
    },
    ytick={0,0.2,0.4,0.6,0.8},
    yticklabels={
        \scriptsize 0,
        \scriptsize 0.2,
        \scriptsize 0.4,
        \scriptsize 0.6,
        \scriptsize 0.8
    },
    tick label style={font=\scriptsize},
    xlabel={\footnotesize (b) Velocity MSE gap},
    xlabel style={yshift=-0.1em},
    ylabel={Mean Velocity MSE},
    ylabel style={
        font=\scriptsize,
        at={(axis description cs:-0.11,0.5)},
        anchor=south
    },
    scaled ticks=false,
    axis line style={line width=0.5pt},
    tick style={line width=0.5pt},
    legend columns=3,
    legend style={
        at={(0.50,1.02)},
        anchor=south,
        fill=none,
        draw=none,
        font=\scriptsize,
        column sep=0.45em
    }
]

\addplot[
    color2,
    mark=o,
    mark size=1.45pt,
    line width=1.0pt,
    mark options={
        solid,
        fill=color2,
        draw=color2
    }
] table[
    x index=0,
    y index=1,
    col sep=space
] {fig/data/Velocity-mse.txt};
\addlegendentry{GenEval}

\addplot[
    color3,
    mark=square,
    mark size=1.35pt,
    line width=1.0pt,
    mark options={
        solid,
        fill=color3,
        draw=color3
    }
] table[
    x index=0,
    y index=2,
    col sep=space
] {fig/data/Velocity-mse.txt};
\addlegendentry{OCR}

\addplot[
    color6,
    mark=diamond,
    mark size=1.45pt,
    line width=1.0pt,
    mark options={
        solid,
        fill=color6,
        draw=color6
    }
] table[
    x index=0,
    y index=3,
    col sep=space
] {fig/data/Velocity-mse.txt};
\addlegendentry{PickScore}

\end{axis}

}
\end{tikzpicture}

%% file: fig/sde_ode.tex
\begin{tikzpicture}
{\footnotesize

\begin{axis}[
    name=plot1,
    width=0.60\linewidth,
    height=0.30\linewidth,
    grid=major,
    grid style={dashed,black!25},
    xmin=35,
    xmax=325,
    ymin=0.62,
    ymax=0.95,
    xtick={40,160,320},
    xticklabels={
        \scriptsize 40,
        \scriptsize 160,
        \scriptsize 320
    },
    ytick={0.65,0.70,0.75,0.80,0.85,0.90,0.95},
    yticklabels={
        \scriptsize 0.65,
        \scriptsize 0.70,
        \scriptsize 0.75,
        \scriptsize 0.80,
        \scriptsize 0.85,
        \scriptsize 0.90,
        \scriptsize 0.95
    },
    tick label style={font=\scriptsize},
    xlabel={\footnotesize Training checkpoint},
    xlabel style={yshift=-0.1em},
    ylabel={GenEval score},
    ylabel style={
        font=\scriptsize,
        at={(axis description cs:-0.11,0.5)},
        anchor=south
    },
    scaled ticks=false,
    axis line style={line width=0.5pt},
    tick style={line width=0.5pt},
    legend columns=4,
    legend style={
        at={(0.50,1.02)},
        anchor=south,
        fill=none,
        draw=none,
        font=\scriptsize,
        column sep=0.45em
    }
]

% =========================================================
% Base model
% =========================================================
\addplot[
    black!55,
    densely dotted,
    line width=1.0pt
] table[
    x=checkpoint,
    y=base,
    col sep=space
] {fig/data/sde_ode_geneval.txt};
\addlegendentry{Base}

% =========================================================
% Task-specific teacher
% =========================================================
\addplot[
    color2,
    dashed,
    line width=1.0pt
] table[
    x=checkpoint,
    y=teacher,
    col sep=space
] {fig/data/sde_ode_geneval.txt};
\addlegendentry{Teacher}

% =========================================================
% ODE sampling
% =========================================================
\addplot[
    color5,
    mark=o,
    mark size=1.45pt,
    line width=1.0pt,
    mark options={
        solid,
        fill=color5,
        draw=color5
    }
] table[
    x=checkpoint,
    y=ode,
    col sep=space
] {fig/data/sde_ode_geneval.txt};
\addlegendentry{ODE}

% =========================================================
% SDE sampling
% =========================================================
\addplot[
    color6,
    mark=square,
    mark size=1.35pt,
    line width=1.0pt,
    mark options={
        solid,
        fill=color6,
        draw=color6
    }
] table[
    x=checkpoint,
    y=sde,
    col sep=space
] {fig/data/sde_ode_geneval.txt};
\addlegendentry{SDE}

\end{axis}

}
\end{tikzpicture}